\documentclass{article}
\usepackage{spconf,amsmath,graphicx}

\usepackage{latexsym}
\usepackage{amsmath}
\usepackage{amsfonts}
\usepackage{bbm}
\usepackage{url}
\usepackage{multirow}
\usepackage{booktabs}
\usepackage{microtype}
\usepackage{hyperref}

\title{GENERATING EMPATHETIC RESPONSES BY LOOKING AHEAD THE USER'S SENTIMENT}
\name{Jamin Shin\textsuperscript{*}\thanks{\noindent\textsuperscript{*}JS and PX contributed equally to this work. A previous version of this work was on arXiv under the name of ``HappyBot"} \qquad Peng Xu\textsuperscript{*} \qquad Andrea Madotto \qquad Pascale Fung}
\address{
Center for Artificial Intelligence Research (CAiRE)\\
The Hong Kong University of Science and Technology\\
\texttt{jayshin.nlp@gmail.com} \qquad \texttt{pasacle@ece.ust.hk}
}
\begin{document}
%
\maketitle
\begin{abstract}
An important aspect of human conversation difficult for machines is conversing with empathy, which is to understand the user's emotion and respond appropriately.
Recent neural conversation models that attempted to generate empathetic responses either focused on conditioning the output to a given emotion, or incorporating the \textit{current} user emotional state.
However, these approaches do not factor in how the \textit{user} would feel \textit{towards} the generated response. 
Hence, in this paper, we propose \textit{Sentiment Look-ahead}, which is a novel perspective for empathy that models the \textit{future} user emotional state. In short, Sentiment Look-ahead is a reward function under a reinforcement learning framework that provides a higher reward to the generative model when the generated utterance improves the user's sentiment. We implement and evaluate three different possible implementations of sentiment look-ahead and empirically show that our proposed approach can generate significantly more empathetic, relevant, and fluent responses than other competitive baselines such as multitask learning.
\end{abstract}
\section{Introduction}

Showing empathy is a core attribute of human conversations as our daily interactions often involve discussions of emotional situations ranging from recent job promotions to even funerals, or about the difficulties of raising children. Hence, it is natural to think that modeling empathy and eliciting it in chatbots are crucial towards bringing them even closer to humans. 

Moreover, there are also practical benefits to incorporating such emotional understanding capabilities. For instance, \cite{martinovski2003breakdown} has reported that addressing the users' emotions reduces the probability of dialogue breakdowns, and \cite{prendinger2005recognizing} has shown that it can enhance overall user satisfaction. In addition, talking about emotional experiences is known to help relieve stress~\cite{zech2005talking} which can also be seen from the large number of users talking about their emotional situations with Microsoft XiaoIce~\cite{zhou2018design}.

As a result, there have been several attempts to model empathy in dialogues. Some initial works include that of~\cite{polzin2000emotion} which manages the dialogue based on user emotions, and creating affective listeners that can respond in terms of both content and affect level~\cite{skowron2010affect}. However, these studies were mainly rule-based systems that are limited in scalability to dataset size and generalizability to different domains and situations. 

\begin{figure}[t]
\includegraphics[width=\columnwidth]{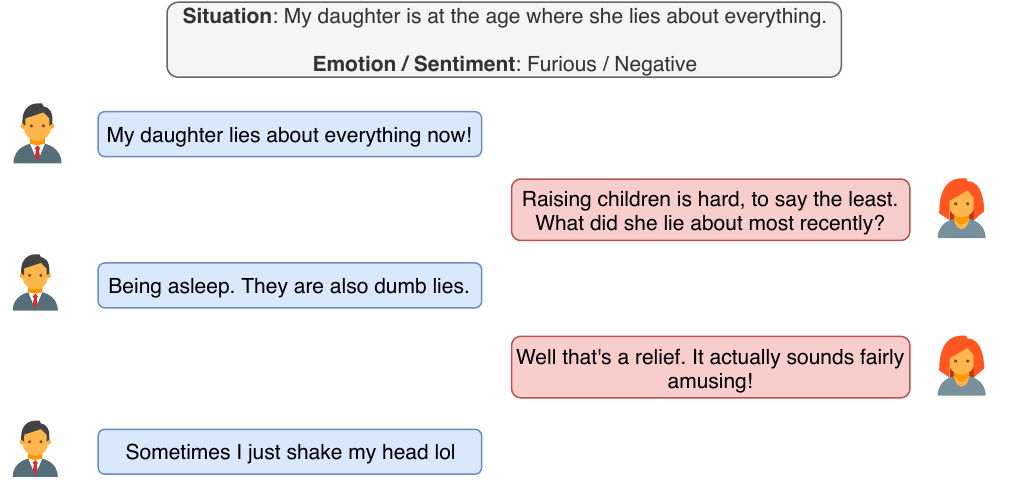}
\caption{Example conversation from Rashkin et al., 2019 \cite{rashkin2019towards}. A speaker (blue) talks about his emotional situation, and the listener (red) is supposed to react empathetically. The task of empathetic dialogue generation aims to model the latter.}
\label{fig:conv}
\end{figure}

\begin{figure*}[ht]
\includegraphics[width=\linewidth]{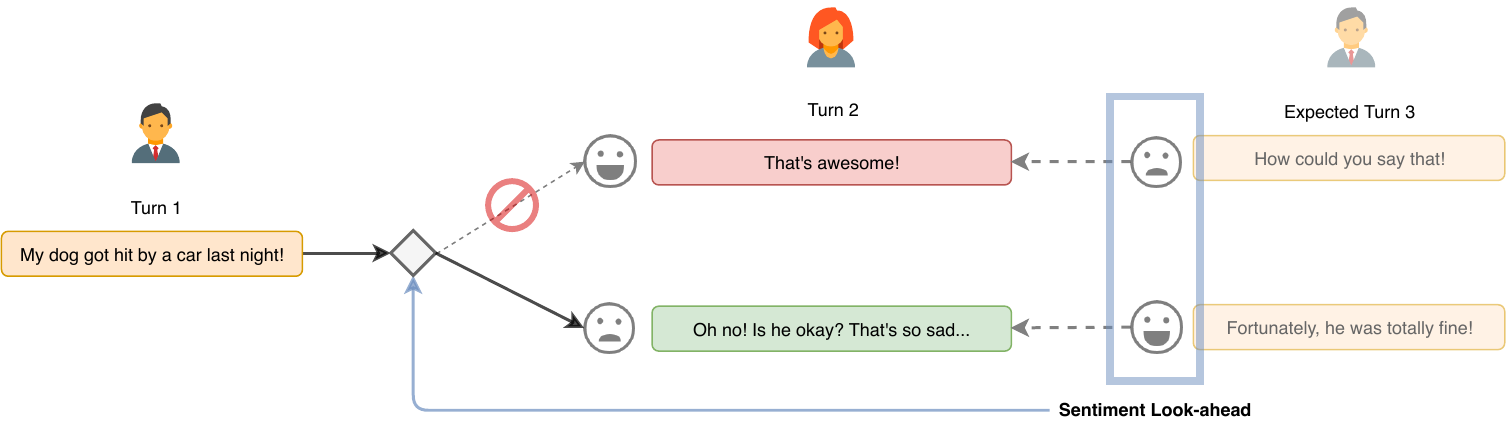}
\caption{Illustration of Sentiment Look-ahead. During conversations, empathetic people consider the impact of their utterances before speaking.}
\label{fig:sentiment_lookahead}
\end{figure*}

Recently, neural conversation models~\cite{vinyals2015neural} have been successful in generating fluent and relevant responses. However, their responses are widely known to promote \textit{dull} and \textit{generic} responses due to the maximum likelihood objective~\cite{li2016diversity,li2016deep} that does not factor in any kind of emotional exchange, or empathy.
Meanwhile, several recent works have tackled the problem of empathetic dialogue response generation, which is understanding the user's emotion and responding appropriately, mainly on two directions. The first line of work has been successful in controlling and conditioning the generated responses to certain sentiments, emotions, and emojis~\cite{hu2017toward,wang2018sentigan,zhou2018mojitalk,zhou2018emotional}. Meanwhile, others have worked on more data-driven approaches by training a model to jointly predict the current emotional state and generate a response~\cite{lubis2018eliciting,rashkin2019towards}. 

While both cases have been successful in generating empathetic and emotional responses to a certain extent, but have neglected some crucial points in empathetic dialogue response generation. 1) The first approach - namely controlled text generation - assumes that the emotion to be conditioned on is given as an input, but we often do not know which emotion is appropriate in order to generate an empathetic response. 2) The latter takes the assumption that by understanding the user's \textit{current} emotion, the model will implicitly learn how to respond empathetically. However, without providing any additional inductive bias, recognizing the current emotional state does not necessarily imply that the model will respond appropriately towards that emotion.

On the other hand, to cope with the above issues, we propose \textit{Sentiment Look-ahead} to directly address the problem of \textit{responding appropriately}. Intuitively, an empathetic person would naturally consider the conversation partner's feelings before speaking and, in turn, trigger a more positive sentiment response from the user. To elaborate, we cast this problem as reinforcement learning in which the reward signal to maximize is the (predicted) sentiment of the \textit{next user turn}.

Finally, we propose three different implementations of the \textit{sentiment look-ahead} reward function. Based on thorough automatic / human evaluations, we show that our approach significantly outperforms other models in 
terms of Empathy, Relevance, and Fluency, verifying the effectiveness of our novel viewpoint about modeling empathy.

\section{Methodology}
Our model has two modules: the policy model and the reward.

\noindent The policy model takes input $x$ and generates a response $\hat{y}$ and the reward model evaluates the generated response with a predefined criteria and updates the policy model accordingly.

\subsection{Notations}
Considering a conversation between a user and the system, we can represent an input dialogue history as an alternating sequence of turns between the two parties as such: $x = [u^1; s^1; u^2; s^2; \cdots; u^T]$, where $T$ is the number of user/system turns and $u^*$ and $s^*$ denote user and system utterances, respectively. The turns in $x$ are flattened as a single sequence of words: $x = [w^1, w^2, \cdots, w^N]$, where $N$ is the number of words in the dialog history. Subsequently, we denote the true output system response $s^{T}$ towards the input $x$ as $y$ and the prediction as $\hat{y}$, where $y$ and $\hat{y}$ are both sequences of words. We also denote $d$ as the dimensionality of the encoder and decoder, and $V$ as the size of the vocabulary.

\subsection{Reward Functions: Sentiment Look-ahead}
\label{sec:reward}
In a nutshell, given the generated system response $s^t$, an arbitrary function $f$ will output a score $R_* \in [0, 1]$ that determines how good the input is, which is then used as the reward for learning the optimal policy. In short, the reward score comes from the next user turn, rather than the currently generated system turn. In the following, we propose and explain three different implementations for sentiment look-ahead reward function and one method for control. We denote function $g(\cdot) \in [0, 1]$ as a pre-trained sentiment classifier that outputs the probability of the input sentence's sentiment being positive.


\noindent \textbf{Forward $R_F$} is the most naive implementation of sentiment look-ahead by directly aiming to predict the next user sentiment towards the generated system response. This reward function outputs the probability of the next user turn sentiment being positive. Hence, $f$ is a GRU trained to predict $g(u^{t+1})$ given $s^{t}$.


\noindent \textbf{Improvement $R_I$} takes a slightly different approach. Because the labels for learning $R_F$ and $g$ are the same 70\% of the time, we hypothesize that $R_F$ will not learn how to predict the next sentiment, but rather the current one.
Instead, we reward our system when the sentiment of the next user turn is predicted to \textit{improve}. Thus, this reward function predicts \textit{whether the sentiment of $u^{t+1}$ improves} compared to the current user turn $u^t$. Hence, $f$ is a GRU trained to predict $\mathbbm{1}(g(u^{t+1}) > g(u^{t}))$ given $s^{t}$.


\noindent \textbf{Simulation $R_S$} is directly given by the sentiment of the \textit{simulated} next user turn $\hat{u}_{t+1}$. As both of the above methods use the \textit{predicted} user reaction instead of the \textit{actual} one, we aim to cope with this issue through user simulation with a pre-trained Seq2Seq model. The sentiment difference $R_S = g(\hat{u}_{t+1}) - g(x)$ is used as the reward.

\noindent \textbf{Current $R_C$} is given by the current sentiment of the input $R_C = g(s^t)$. This model is used as control in order to isolate the contributions of sentiment look-ahead models.

\subsection{Policy Model: Seq2Seq with Attention}
\label{sec:policy}
The Policy Model is a Seq2Seq model~\cite{sutskever2014sequence} based on GRU~\cite{chung2014empirical} and dot product attention~\cite{luong2015effective}. 
To elaborate, the encoder BiGRU takes the flattened input $x$ and generates the encoder hidden states $\mathbf{h}_{enc} = [h^1_{enc}, h^2_{enc}, \cdots h^N_{enc}] \in \mathbb{R}^{d\times N}$. 
For each decoding step $t$, the decoder generates the decoder hidden state $h^t_{dec} \in \mathbb{R}^{1 \times d}$ given the previous decoder state $h^{t-1}_{dec}$, the previous target token $y^{t-1}$, and the context vector using attention $c^t$:
\begin{align}
    \mathbf{h}_{enc} &= \text{BiGRU}(x) \\
    h^t_{dec} &= \text{GRU}(h^{t-1}_{dec}, y^{t-1}, c^{t-1}) \\
    a_t &= \text{softmax}(h^t_{dec} \cdot \mathbf{h}_{enc}) \\
    c^t &= \sum_{i=1}^N a_t^i h^i_{enc}
\end{align}
, where $y^0$ is $<$SOS$>$ token, $h^0_{dec} = \overrightarrow{0}$, and $c^0 = \overrightarrow{0}$. 

The decoder hidden state and context vector is then used for generating a conditional probability distribution the next token $\hat{y}_t$ is sampled from:
\begin{gather}
    h_t = [c^t; h^t_{dec}] \\
    h_t^* = \text{tanh}(h_t \cdot \mathbf{W}_r) \\
    \hat{y}_t \sim p_\theta(y~|~y^{t-1},~h^{t-1}_{dec}) = \text{softmax} ( h_t^* \cdot \mathbf{W}_{out} / \tau) \label{eq:prob}
\end{gather}
, where $\mathbf{W}_r \in \mathbb{R}^{2d \times d}$ is a trainable parameter that reduces the hidden state dimensionality, $\mathbf{W}_{out} \in \mathbb{R}^{d \times V}$ is a trainable matrix that maps the reduced state to the output vocabulary space, $\tau$ refers to softmax temperature, and $\theta$ is parametrized by $\{\text{BiGRU}, \text{GRU}, \mathbf{W}_r, \mathbf{W}_{out} \}$.

\subsection{Policy Learning} 
We follow the MIXER algorithm~\cite{ranzato2015sequence} which basically pre-trains the Policy Model and incrementally updates it with REINFORCE~\cite{williams1992simple}. First, we optimize the following Maximum Likelihood Estimation (MLE) loss function $\mathcal{L}_\text{MLE}$ to pre-train the Policy Model:
\begin{align}
    \mathcal{L}_\text{MLE} 
    &= - \log p(y)
    = - \log p(y^1, \cdots, y^M) \\
    &= - \frac{1}{M} \sum\nolimits_{t=1}^M \log p(y^t~|~y^1, \cdots, y^{t-1}) \label{eq:mle}
\end{align}
, where $M$ is the number of words to generate and $p(y^t~|~\cdot)$ is parametrized by Equation~\ref{eq:prob}.

We then use REINFORCE which defines the loss function $\mathcal{L}_{RL}$ as the negative expected reward:
\begin{gather}
    \hat{R_t} = h^t_{dec} \cdot \mathbf{W_b} \\
    \mathcal{L}_{\text{RL}} = - \frac{1}{M} \sum_{t=1}^M  (R_* - \hat{R_t}) \log p(\hat{y}_t~|~\hat{y}_1,\cdots,\hat{y}_{t-1})  \label{eq:rl}
\end{gather}
, where $\mathbf{W_b} \in \mathbb{R}^{d \times 1}$ is a trainable parameter, $\hat{R}_t$ is the baseline reward, and $R_*$ is an arbitrary reward. The baseline reward is used in order to reduce the variance of the actual reward and is a linear model trained by minimizing the Mean Squared Error (MSE) loss between $R_*$ and $\hat{R}_t$:
\begin{equation}
    \min_{\mathbf{W_b}} \frac{1}{T} \sum\nolimits_{t=1}^T | R_* - \hat{R_t} |^2
\end{equation}

\subsection{Hybrid Training}
As the reward function from REINFORCE may deteriorate the model from generating fluent and relevant response, we follow~\cite{paulus2018deep,zhou2018mojitalk} to stabilize the training process by employing a mixed learning objective which augments the policy learning objective after pre-training with $\mathcal{L}_\text{MLE}$ as such:
\begin{equation}
    \mathcal{L} = \lambda \mathcal{L}_\text{RL} + (1 - \lambda) \mathcal{L}_\text{MLE}
\end{equation}
, where $\lambda$ is a hyper-parameter that interpolates each loss.

\begin{table*}[ht!]
\centering
\resizebox{0.9\textwidth}{!}{%
\begin{tabular}{@{}cccccc@{}}
\toprule
\multicolumn{6}{c}{\textbf{Automatic Evaluation Results}}                                         \\ \midrule
            & Sentence Length & Distinct 1-gram & Distinct 2-gram & Distinct 3-gram & Average BLEU   \\ \midrule
Human       & 14.463          & 0.065           & 0.386           & 0.725           & -             \\ \midrule
Seq2Seq     & 11.494 $\pm$ 0.0281        & 0.019 $\pm$ 0.0003          & 0.115 $\pm$ 0.0003          & 0.263 $\pm$ 0.0014          & \textbf{6.558} $\pm$ 0.0520        \\
MultiSeq    & \textbf{13.264} $\pm$ 0.0298 & 0.015 $\pm$ 0.0003          & 0.102 $\pm$ 0.0007          & 0.246 $\pm$ 0.0012          & 6.250 $\pm$ 0.0433        \\ \midrule
Current     & 12.799 $\pm$ 0.0928         & 0.015 $\pm$ 0.0004          & 0.100 $\pm$ 0.0009          & 0.240 $\pm$ 0.0023          & 6.275 $\pm$ 0.0250        \\ \midrule
Forward     & 13.177 $\pm$ 0.0122         & 0.020 $\pm$ 0.0004          & 0.123 $\pm$ 0.0004          & 0.285 $\pm$ 0.0004          & 6.292 $\pm$ 0.0289        \\
Improvement & 13.098 $\pm$ 0.0368         & \textbf{0.026} $\pm$ 0.0004  & \textbf{0.170} $\pm$ 0.0013  & \textbf{0.379} $\pm$ 0.0025  & 6.317 $\pm$ 0.0144        \\
Simulation  & 12.154 $\pm$ 0.0432         & 0.011 $\pm$ 0.0001          & 0.080 $\pm$ 0.0004          & 0.202 $\pm$ 0.0015          & 6.192 $\pm$ 0.0629        \\ \bottomrule
\end{tabular}
}
\caption{Automatic metrics measuring fluency of generated responses. \textit{Improvement} model significantly outperforms others in all Distinct N-gram measures. It also achieves the second highest in Sentence Length and Avereage BLEU score.}
\label{tab:auto_fluency}
\end{table*}

\section{Experiment Setting}
We mainly conduct our experiments and evauations on the EmpatheticDialogue~\cite{rashkin2019towards}~dataset, which consists of 25k conversations between a \textit{Speaker} and a \textit{Listener} about a given emotional situation. The dataset provides evenly distributed 32 emotion labels that we map down to positive/negative sentiments. To continue, in order to promote diversity of the response generation, we pre-train the policy model on a combined dataset of EmpatheticDialogue, DailyDialog~\cite{li2017dailydialog}, and PersonaChat~\cite{zhang2018personalizing}.

Before training the sentiment look-ahead reward functions, we first have to train the sentiment classifier $g(\cdot)$ from Section~\ref{sec:reward}, which is also $R_C$. For training the sentiment classifier, we fine-tune the pre-trained BERT~\cite{devlin2019bert} on SST-2~\cite{socherRNTN} dataset and the situation texts from EmpatheticDialogue by mapping emotions to binary sentiment labels as mentioned earlier. We then use this to label the sentiments of each turn, generate forward sentiment and sentiment improvement labels, and train the GRU-based reward models $R_F$ and $R_I$.

We use the pre-trained BERT \textit{base}~\cite{devlin2019bert} model, 300 dimensional pre-trained FastText\footnote{\url{https://fasttext.cc/}} word embeddings, hidden size $d=300$ for GRU cells, and tie the embedding weights of the decoder's output layer as in~\cite{merity2018regularizing}. We use learning rates of [$1\mathrm{e}{-3}$, $1\mathrm{e}{-4}$] and hybrid training ratio $\lambda$ of 0.5. Finally, we use top-$k$ sampling~\cite{fan2018hierarchical} using $k=40$ along with softmax temperature $\tau$ of 0.5.\footnote{Our source code is released at \\ \href{https://github.com/HLTCHKUST/sentiment-lookahead}{\url{https://github.com/HLTCHKUST/sentiment-lookahead}}}

\subsection{Evaluation}

\noindent \textbf{Automatic Metrics}
We evaluate 4 metrics that represent how fluent and relevant the generated responses are:

\begin{itemize}
    \item \textbf{Sentence Length}: Measure average length of generated utterances.
    \item \textbf{Distinct $\{1,2,3\}$-grams}: Count percentage of unique $\{1,2,3\}$-grams as in~\cite{li2016diversity}.
    \item \textbf{Average BLEU}: Average of BLEU- $\{1,2,3,4\}$ as in~\cite{rashkin2019towards}.
    \item \textbf{Bag of Words Embedding Similarity}: Three different types of cosine similarities (\textbf{Extrema}, \textbf{Average}, \textbf{Greedy}) of Bag-of-Words Embedding between generation and target as in~\cite{liu2016not}.
\end{itemize}

\noindent \textbf{Human Evaluation}
\label{subsub:human_eval}
We modify the human evaluation method from~\cite{rashkin2019towards} as 5-point scale does not directly compare between models. Instead of 5-point ratings, we conduct Multiple Choice Testing against Baselines and task 5 human judges with 100 dialogue samples. We ask them to choose among 6 different models the best responses for each criterion (\textbf{Fluency}, \textbf{Relevance}, and \textbf{Empathy}). We also gave them options to choose 'Tie - All' and 'Tie - None' which indicate all answers were good or all answers were bad.

\section{Results \& Discussion}
\subsection{Discussion of Automatic Evaluation Results}
Because of the stochastic nature of using top-k sampling~\cite{fan2018hierarchical} as a decoding strategy, we run each evaluation 3 times and report the mean and standard deviation.

From Table~\ref{tab:auto_fluency}, it is clearly visible that \textit{Improvement} model significantly outperforms other proposed models and baselines in terms of Distinct N-grams. This indicates that this model can generate more diverse sentences than the others. \textit{Improvement} model having the second highest Sentence Length and Average BLEU also points out that our \textit{Improvement} model generates more fluent utterances.

\begin{table}[t]
\centering
\resizebox{0.47\textwidth}{!}{%
\begin{tabular}{@{}cccc@{}}
\toprule
\multicolumn{4}{c}{\textbf{Bag of Words Embedding Similarity}}     \\ \midrule
            & Extrema         & Average         & Greedy          \\ \midrule
Seq2Seq     & 0.908 $\pm$ 0.0015        & 0.961 $\pm$ 0.0005        & 0.949 $\pm$ 0.0007        \\
MultiSeq    & 0.904 $\pm$ 0.0013        & 0.962 $\pm$ 0.0011        & 0.948 $\pm$ 0.0003        \\ \midrule
Current     & 0.903 $\pm$ 0.0022        & 0.961 $\pm$ 0.0009        & 0.948 $\pm$ 0.0004        \\ \midrule
Forward     & 0.901 $\pm$ 0.0011        & 0.961 $\pm$ 0.0006        & 0.947 $\pm$ 0.0014        \\
Improvement & \textbf{0.919} $\pm$ 0.0008        & \textbf{0.966} $\pm$ 0.0003        & \textbf{0.952} $\pm$ 0.0006 \\
Simulation  & \textbf{0.919} $\pm$ 0.0020 & \textbf{0.966} $\pm$ 0.0011  & 0.949 $\pm$ 0.0003          \\ \bottomrule
\end{tabular}
}
\caption{Automatic metrics for measuring relevance of generated responses using Bag-of-Words Embedding Cosine Similarity with human reference. \textit{Improvement} and \textit{Simulation} models show highest scores in all three types.}
\label{tab:auto_relevance}
\end{table}

We evaluate how relevant the generated sentences are using three different types of Bag-of-Words Embedding Similarity as in~\cite{liu2016not}, namely Extrema, Average, and Greedy. For all metrics, we use FastText embeddings and measure the cosine similarity between generated utterances and human responses in order to check how semantically close our responses are to the references. For all three metrics, Table~\ref{tab:auto_relevance} shows us that \textit{Improvement} and \textit{Simulation} achieves higher scores than other baselines, which indicates that these responses could be considered more semantically closer to the human responses.

\begin{table}[t]
\centering
\resizebox{0.4\textwidth}{!}{
\begin{tabular}{@{}cccc@{}}
\toprule
\multicolumn{4}{c}{\textbf{Multiple Choice Testing against Baselines}} \\ \midrule
\textbf{}        & Empathy         & Relevance       & Fluency        \\ \midrule
Seq2Seq          & 0.11            & 0.14            & 0.05           \\
MultiSeq         & 0.12            & 0.09            & 0.03           \\
Current          & 0.11            & 0.12            & 0.03           \\
Forward          & 0.09            & 0.1             & 0.07           \\
Improvement      & \textbf{0.24}   & \textbf{0.26}   & \textbf{0.12}  \\
Simulation       & 0.07            & 0.04            & 0.01           \\
Tie - All        & 0.08            & 0.06            & 0.68           \\
Tie - None       & 0.18            & 0.19            & 0.01           \\ \midrule
Total            & 1.0             & 1.0             & 1.0            \\ \midrule
\textbf{Kappa}   & 0.27            & 0.2             & 0.08           \\ \bottomrule
\\
\toprule
\multicolumn{4}{c}{\textbf{A/B Testing against Human}} \\ \midrule
\textbf{}           & Empathy      & Relevance      & Fluency     \\ \midrule
Improvement         & 0.14         & 0.05           & 0.05        \\
Human               & \textbf{0.4}          & 0.42           & 0.36        \\
Tie - All           & 0.37         & \textbf{0.44}           & \textbf{0.52}        \\
Tie - None          & 0.09         & 0.09           & 0.07        \\ \midrule
Total               & 1.0          & 1.0            & 1.0         \\ \midrule
\textbf{Kappa}      & 0.24         & 0.31           & 0.07        \\ \bottomrule
\end{tabular}
}
\caption{
\textit{Improvement} model clearly outperforms all other models in all three tested categories, where Kappa is annotator agreement score.
For A/B testing, \textit{Human} outperforms \textit{Improvement} mode, but the ratios of \textit{Tie - All} are also significantly high.}
\label{tab:mc_test}
\end{table}

\subsection{Discussion of Human Evaluation Results}

We conduct two types of multiple-choice human surveys: one against other models and another testing the best model against human responses. 

\noindent \textbf{Multiple-choice Testing}
Note that for Table~\ref{tab:mc_test}, \textit{Kappa} indicates Fleiss-Kappa multi-annotator agreement score~\cite{fleiss1971measuring}. Heuristically, a score between 0.2 and 0.4 is within range of ``fair agreement.''
We chose to do a multiple choice test so that we could conduct a more direct comparison with other models and find the best, whereas 5-point scale as in~\cite{rashkin2019towards} would result in an independent set of results that are not directly comparable. From Table~\ref{tab:mc_test}, we could easily see that \textit{Improvement} model results the highest in all three categories, which is consistent with the automatic results. For Fluency, it is clear that the vast majority of the human judges selected \textit{Tie All}. This is because most answers were often grammatically accurate.


\noindent \textbf{Humans vs Improvement}
We then conducted an A/B test between the best performing model (\textit{Improvement}) against human responses in order to see to what extent we can compete with humans. Unsurprisingly, as shown in Table 6, human responses are chosen approximately 40\% of the time for all three categories. However, it is notable that \textit{Tie - All} was chosen similarly as well, which means that the judges were not sure which one was better as both were good responses.

\begin{table}[t!]
\centering
\resizebox{0.45\textwidth}{!}{%
\begin{tabular}{@{}l@{}}
\toprule
\textbf{Dialogue History} \\
\textit{SPEAKER}: Not all of my high-school memories are fond\\
ones. Like the time I drank too much and passed out\\
at a party. \\ 
\midrule 
\textbf{Seq2Seq} \\
\textit{LISTENER}: I'm sorry to hear that. I'm sure you'll find\\
a nice place to be a party.\\
\midrule 
\textbf{MultiSeq} \\
\textit{LISTENER}: I know how you feel. I don't have kids, but \\
I don't blame you.\\
\midrule 
\textbf{Current} \\
\textit{LISTENER}: That is awesome. I bet you were so proud. \\
\midrule 
\textbf{Forward} \\
\textit{LISTENER}: I love surprises, especially when you're a kid.\\
I know it has to feel really bad about it.\\
\midrule
\textbf{Improvement} \\
\textit{LISTENER}: That's too bad. What was your favorite part?\\ 
\midrule
\textbf{Simulation} \\
\textit{LISTENER}: I'm so sorry to hear that. Did you have fun?\\
\bottomrule 
\end{tabular}
}
\caption{ Generated dialogue responses from different models. Our method (\textit{Improvement}) produces the most empathetic and relevant response.}
\label{tab:generations}
\end{table}

\subsection{Generation Examples}
We then look at some generation examples to verify the insights from previous discussions and discoveries made in the above correlation analysis. From the first dialogue in Table~\ref{tab:generations}, we can see that all responses show positive sentiments and are fluent in terms of grammar, but \textit{Improvement} exhibits the longest and most relevant response by copying the word ``beach'' from the input. Furthermore, for the second dialogue example, \textit{Seq2Seq} is the only model to directly address ``party", but it is not really relevant or fluent. On the other hand, \textit{Improvement} not only properly addresses the SPEAKER's sentiment, but also asks what were the good memories instead of further discussing the bad ones.

\section{Conclusion}
In this paper, we propose a novel perspective on empathy, \textit{sentiment look-ahead}, as the key to generating empathetic responses. We implement and train three different sentiment look-ahead reward functions to model how the user would feel towards a generated dialogue response, and use such to encourage more empathetic responses. The empirical results on both automatic and human evaluations in terms of Empathy, Fluency, and Relevance confirm that our proposed approach is an effective way to generate more empathetic responses compared to other models that condition on current user emotions.

\bibliographystyle{IEEEbib}
\bibliography{strings,refs}

\end{document}